\documentclass[10pt,a4paper]{article}
\usepackage[margin=0.82in]{geometry}
\usepackage[T1]{fontenc}
\usepackage{lmodern,microtype}
\usepackage{amsmath,amssymb,bm,mathtools}
\usepackage{booktabs,tabularx,array,multirow}
\usepackage{graphicx}
\usepackage{tikz}
\usetikzlibrary{arrows.meta,positioning,calc,fit,backgrounds,decorations.pathreplacing}
\usepackage{xcolor}
\usepackage{enumitem}
\usepackage[round,authoryear]{natbib}
\usepackage[hidelinks]{hyperref}
\usepackage{caption}
\usepackage{float}
\usepackage{needspace}
\usepackage{fancyhdr}
\definecolor{ink}{HTML}{21374B}
\definecolor{teal}{HTML}{137B73}
\definecolor{blue}{HTML}{3D69A8}
\definecolor{ochre}{HTML}{A57238}
\definecolor{purple}{HTML}{795E96}
\definecolor{light}{HTML}{F4F7F9}
\definecolor{mute}{HTML}{5C6875}
\hypersetup{pdftitle={How Generative Models Generate: A Concise Guide to the Generative AI Landscape},pdfauthor={Tianhua Chen},pdfsubject={Concise generation-centred technical tutorial and conceptual synthesis}}
\setlist[itemize]{topsep=3pt,itemsep=2pt,leftmargin=*}
\setlist[enumerate]{topsep=3pt,itemsep=2pt,leftmargin=*}
\newcommand{\E}{\mathbb E}

\tikzset{nodebox/.style={draw,rounded corners=2.4pt,align=center,minimum height=8mm,inner sep=4pt,font=\footnotesize},
   arrow/.style={-{Latex[length=2.1mm]},line width=0.80pt},
   dottedarrow/.style={-{Latex[length=1.9mm]},dashed,line width=.65pt},
   sample/.style={draw=teal!80,fill=teal!7},
   cond/.style={draw=blue!75,fill=blue!7},
   evolve/.style={draw=ochre!80,fill=ochre!8},
   learn/.style={draw=purple!70,fill=purple!7}}
   
\title{\vspace{-0.8em}\textbf{How Generative Models Generate:
A Concise Guide to the Generative AI Landscape}}
\author{Tianhua Chen\\
Department of Computer Science, \\University of Huddersfield, UK\\
\texttt{T.Chen@hud.ac.uk}}
\date{} % Suppress the default date in the title block

\begin{document}
\maketitle
\vspace{-1.5em}
\begin{abstract}
The rapidly expanding generative AI landscape brings together methods that differ
in architecture, learning objective, mathematical representation and
sampling procedure. A concise, selectively organised technical guide can
provide an accessible starting point for developing a coherent understanding
of this increasingly complex landscape. This article takes a
generation-centred approach, connecting how models produce new observations
with the modelling and learning principles that make generation possible.
We develop a two-level perspective: first, distinguishing three recurring
generation routes---direct transformation or decoding, sequential conditional
construction, and iterative updating or transport; and second, examining
the modelling and learning principles underlying these routes.
Representative approaches are introduced through their key components, learning objectives and sampling procedures. We then examine their broader connections, including how different architectures and generation mechanisms can overlap or be combined. By bringing together essential
technical explanations within a compact conceptual framework, the article
aims to help readers develop a working understanding of unfamiliar
approaches, recognise connections among models they have encountered,
and navigate different approaches within the broader generative AI landscape.
\end{abstract}

\vspace{1pt}

\section{Understanding generative AI through generation}
\label{sec:intro}

How does a modern generative model produce a new observation?
This simple question provides an entry point into a rapidly expanding
landscape in which familiar names describe different aspects of a
system. A Transformer is a neural architecture,
autoregression is a conditional modelling principle, and
flow matching specifies a way of learning a velocity field.
Connecting what is learned with what happens when a new sample is
requested makes these distinctions easier to understand.

Navigating this evolving landscape can be challenging in different
ways. Readers with a background in mathematics or machine learning
may not have studied generative modelling systematically; those
familiar with a few prominent approaches may find unfamiliar ones
difficult to understand; and encountering many model names does not
necessarily reveal how their underlying mechanisms relate.
A wealth of existing resources addresses different needs.
Research papers introduce individual methods, while specialist
tutorials provide deeper explanations of particular approaches,
such as normalising flows \citep{papamakarios2021} and diffusion
models \citep{yang2024diffusion}. Comprehensive comparative surveys
examine established model families and their connections
\citep{bondtaylor2022}, while more recent surveys cover
architectures, variants, theoretical aspects and applications
\citep{almuammar2026,tera2026}.

These resources are valuable for their respective purposes.
However, readers seeking to grasp the essential mechanisms of
different approaches and understand how they fit together may
find it difficult to develop a coherent overall picture from
individual treatments or extensive surveys. A concise,
selectively organised guide can offer a more accessible starting
point without requiring readers to navigate the full breadth
and depth of the literature. This article takes that approach:
it explains enough of the principal components and workings
of representative models to establish a meaningful technical
understanding, while connecting them through a common
generation-centred perspective. The aim is to help readers
develop a working understanding of unfamiliar approaches,
recognise connections among models they have encountered,
and navigate the wider generative AI landscape,
rather than provide either a from-scratch tutorial or an
exhaustive survey.

Generative modelling seeks to capture the statistical structure
of observed data in a form that enables new, representative
samples to be produced. However, the observations themselves
do not provide a procedure for drawing fresh independent samples
from the underlying distribution. Different generative models
address this challenge by learning different mathematical
information and using it through different generation procedures.

Consider image synthesis, where a generative adversarial
network can transform a random latent input directly
into an image \citep{goodfellow2014}, while diffusion and flow-matching models
commonly generate images by progressively updating an
initial noise sample \citep{ho2020,lipman2023}. In text generation, autoregressive
models typically construct an observation one token at
a time, conditioning each new token on those previously
generated \citep{radford2019}. These examples illustrate different ways in
which learned information enables generation, rather
than distinctions imposed by the data modality itself.
Understanding these underlying generation mechanisms,
together with the modelling and learning principles
that support them, provides a useful starting point
for navigating the wider generative AI landscape.

We therefore organise the generative AI landscape through two connected levels. The first concerns how a new
observation is produced at generation time. In direct
generation, sampled inputs are decoded or transformed into a complete
observation; in sequential generation, an observation is
constructed progressively through conditional draws; and in
iterative generation, an existing state is repeatedly
updated or transported. The second level concerns the modelling
and learning principles that make these generation mechanisms
possible. Depending on the method, training may establish a decoder,
generator, conditional distribution, noise predictor, score, velocity
field or energy function. These learned quantities play different
roles in generation, and some require additional sampling or numerical
procedures to produce observations. Importantly, the three routes
are not mutually exclusive model classes: the same architecture can
support different generation mechanisms, and complete systems can
combine or nest multiple routes.

\begin{figure}[!htbp]
\centering
\resizebox{\linewidth}{!}{% High-level overview of two organising levels.
% More detail appears in the route-specific figures.

\begin{tikzpicture}[x=1cm,y=1cm,>=Latex,font=\sffamily]

\def\wide{5.0}
\def\gap{.42}

% --------------------------------------------------
% LEVEL 1: GENERATION ROUTES
% --------------------------------------------------

\node[
    font=\bfseries\scriptsize,
    text=mute,
    anchor=west
] at (0,5.12)
{LEVEL 1: HOW IS A NEW OBSERVATION GENERATED?};

\foreach \left/\accent/\heading/\description in {
   0/teal/{DIRECT}/{Produce a complete\\observation by mapping or decoding},
   5.42/blue/{SEQUENTIAL}/{Build an observation\\component by component},
   10.84/ochre/{ITERATIVE}/{Repeatedly update\\an evolving sample}
}
{
   \path[
       draw=\accent!80,
       fill=\accent!11,
       rounded corners=4pt
   ] (\left,3.82) rectangle ++(5.0,1.10);

   \node[
       font=\bfseries\footnotesize,
       text=\accent
   ] at (\left+2.5,4.65) {\heading};

   \node[
       font=\scriptsize,
       text=ink,
       align=center,
       text width=4.7cm
   ] at (\left+2.5,4.15) {\description};
}

% --------------------------------------------------
% LEVEL 2: MODELLING AND LEARNING PRINCIPLES
% --------------------------------------------------

\node[
    font=\bfseries\scriptsize,
    text=mute,
    anchor=west
] at (0,3.62)
{LEVEL 2: MODELLING AND LEARNING PRINCIPLES};

% Direct generation
\path[
    draw=teal!55,
    fill=teal!3,
    rounded corners=4pt
] (0,1.64) rectangle ++(5.0,1.65);

% Sequential generation
\path[
    draw=blue!55,
    fill=blue!3,
    rounded corners=4pt
] (5.42,1.64) rectangle ++(5.0,1.65);

% Iterative generation
\path[
    draw=ochre!55,
    fill=ochre!3,
    rounded corners=4pt
] (10.84,1.64) rectangle ++(5.0,1.65);

\node[
    font=\footnotesize,
    align=center,
    text width=4.75cm
] at (2.50,2.47)
{
    Probabilistic latent variables\\[3pt]
    Adversarial learning\\[3pt]
    Finite-layer normalising flows
};

\node[
    font=\footnotesize,
    align=center,
    text width=4.75cm
] at (7.92,2.47)
{
    Autoregressive modelling
};

\node[
    font=\footnotesize,
    align=center,
    text width=4.75cm
] at (13.34,2.47)
{
    Denoising and score learning\\[3pt]
    Velocity-based transport\\[3pt]
    Energy-guided sampling
};

% --------------------------------------------------
% GENERATION AFTER LEARNING
% --------------------------------------------------

\node[
    font=\bfseries\scriptsize,
    text=mute,
    anchor=west
] at (0,1.17)
{GENERATION AFTER LEARNING (PARAMETERS FIXED)};

\foreach \left/\accent/\description in {
   0/teal/{Draw $z$ $\to$ generate complete $x$},
   5.42/blue/{Sample $x_i\mid x_{<i}$ $\to$ append},
   10.84/ochre/{Draw initial state $\to$ repeated updates}
}
{
   \path[
       draw=\accent!50,
       fill=\accent!9,
       rounded corners=3pt
   ] (\left,.16) rectangle ++(5.0,.81);

   \node[
       font=\footnotesize,
       align=center,
       text width=4.65cm
   ] at (\left+2.5,.56) {\description};
}

\end{tikzpicture}}
\caption{\textbf{Two-level map of generative AI: generation routes and representative modelling principles.}}
\label{fig:ecosystem}
\end{figure}
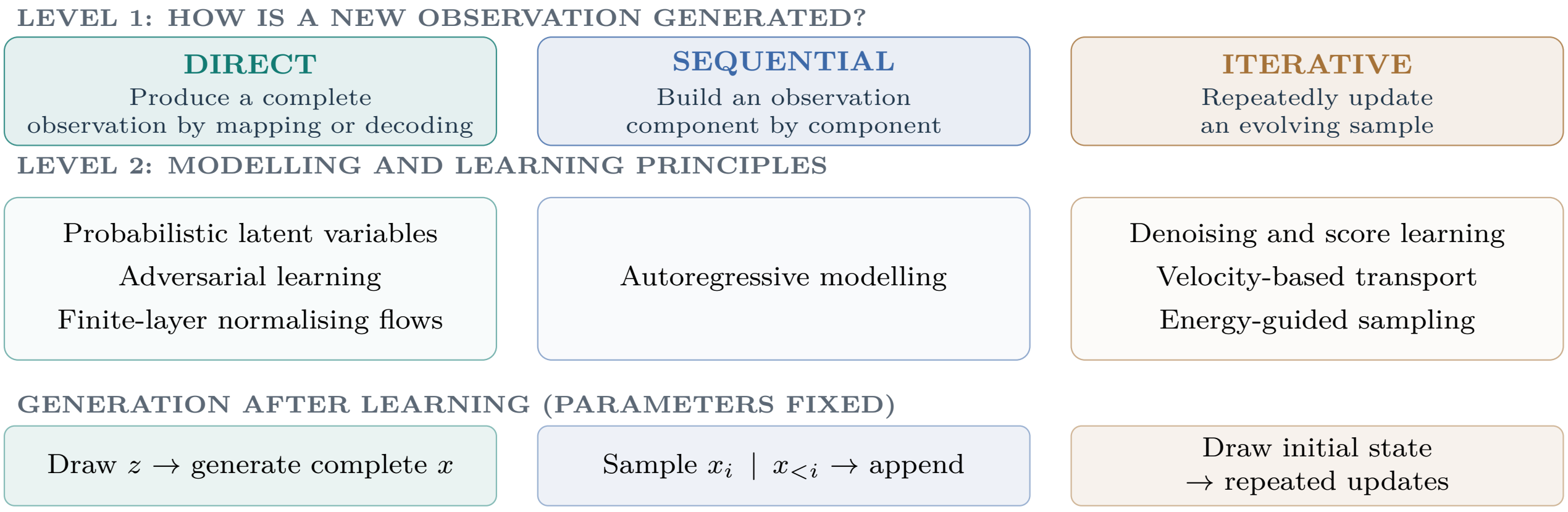

Figure~\ref{fig:ecosystem} provides a high-level map of this
two-level organisation. The remainder of the paper develops
the framework by introducing representative models and explaining
their essential components, learning principles and generation
procedures. Sections~\ref{sec:direct}--\ref{sec:iter} examine the
three generation routes, supported by route-specific illustrations;
Section~\ref{sec:compare} connects the learned quantities and sampling
operations across model families, including how architectures and
generation mechanisms can overlap or be combined. Together, these explanations aim to provide a concise but technically meaningful map of generative AI
modelling.

% Principal conceptual figure in the merged Introduction (Section 1).

\section{Direct generation: from sampled inputs to observations}
\label{sec:direct}

Direct generation commonly begins with randomness drawn from
a simple distribution that is easy to sample from, such as
a standard Gaussian. The central idea is to learn a mechanism
that transforms this randomness into complete observations
whose statistical structure resembles that of the training data.
For a deterministic generator, this takes the form
\begin{equation}\label{eq:direct-general}
 z\sim p_Z(z),\qquad x=G_\theta(z).
\end{equation}
The learned mapping may contain many neural-network layers or
successive transformations. What makes generation direct
is that a complete observation is produced through the learned generative mechanism,
rather than constructed component by component or progressively
refined through a separate iterative sampling procedure.

The main learning challenge is therefore to establish a mapping
or conditional mechanism that turns accessible randomness into
observations with the required statistical structure. Three
influential approaches illustrate different ways to achieve this:
probabilistic latent-variable modelling, adversarial learning and
finite-layer invertible transformations. They share a broad
generation route but differ in their learning objectives and
learned quantities.

\subsection{Probabilistic latent-variable modelling: from classical models to VAEs}
\label{sec:latent-direct}

Probabilistic latent-variable models explain observations through unobserved factors. They specify a law for the hidden variable and a conditional observation model:
\begin{equation}\label{eq:latent-general}
 z\sim p_\eta(z),\qquad x\sim p_\theta(x\mid z),\qquad
 p_{\eta,\theta}(x)=\int p_\theta(x\mid z)p_\eta(z)\,\mathrm dz.
\end{equation}
The prior $p_\eta(z)$ determines how latent values are selected, and $p_\theta(x\mid z)$ describes possible observations given each latent value. For discrete $z$, the integral becomes a sum. Generation needs only the two draws, not an evaluation of the marginal integral.

Learning is harder because the latent values explaining each training observation are not observed. Maximum marginal likelihood seeks to increase $\sum_{i=1}^{N}\log p_{\eta,\theta}(x_i)$, while posterior inference concerns $p_{\eta,\theta}(z\mid x)$. This posterior is determined by the generative parameters $\eta$ and $\theta$, but may be difficult to evaluate. The classical expectation--maximisation (EM) algorithm alternates between obtaining posterior information under the current model (E-step) and updating generative parameters using the expected complete-data log-likelihood (M-step) \citep{dempster1977}. Gaussian mixture models (GMMs) and probabilistic PCA (PPCA) are classical examples that can be fitted using EM \citep{tipping1999}. In more complex models, especially those with flexible nonlinear relationships, evaluating the posterior or marginal likelihood may become intractable. Variational methods instead introduce a separately parameterised approximate posterior, for example $q_\phi(z\mid x)$, and optimise a tractable lower bound, as demonstrated shortly.

Among modern generative AI methods, variational autoencoders (VAEs), introduced in the early 2010s, are a prominent neural example of probabilistic latent-variable modelling \citep{kingma2014}. A VAE uses an encoder--decoder architecture: the encoder infers a latent representation for an observed data point, while the decoder turns latent values into a distribution over possible observations. Like PPCA, the model explains observations through continuous latent variables, but neural networks allow the relationship to be nonlinear. More precisely, the encoder $q_\phi(z\mid x)$ approximates the generally intractable posterior $p_\theta(z\mid x)$, while the decoder $p_\theta(x\mid z)$ provides the conditional generative model. Rather than returning just a fixed PCA code, a common encoder predicts a Gaussian distribution for each input. A prior $p_Z(z)$, often $\mathcal N(0,I)$, supplies latent values during unconditional generation.

A VAE jointly learns its encoder and decoder by maximising the evidence lower bound (ELBO), a standard variational objective for latent-variable models:
\begin{equation}\label{eq:elbo}
\log p_\theta(x)\ge
\underbrace{\mathbb E_{q_\phi(z\mid x)}[\log p_\theta(x\mid z)]}_{\text{reconstruction}}
-\underbrace{D_{\mathrm{KL}}\!\left(q_\phi(z\mid x)\,\Vert\,p_Z(z)\right)}_{\text{prior regularisation}}.
\end{equation}
The reconstruction term rewards a decoder that explains $x$ using latent values drawn from its approximate posterior; for a fixed-variance Gaussian decoder, this corresponds to squared reconstruction error up to scaling and constants. The KL term discourages excessive deviation of that posterior from the chosen prior. It helps connect training-time latent inference with generation-time sampling: the generator will later draw $z$ from $p_Z$, not from an encoder conditioned on a particular observation. The terms must remain balanced; if the encoder returns the prior regardless of $x$, the latent variable may carry little information for reconstruction. %The KL penalty encourages prior compatibility, but does not by itself guarantee a useful or completely covered latent representation.

In practice, a common Gaussian encoder learns to predict the latent mean $\mu_\phi(x)$ and scale $\sigma_\phi(x)$ (hence variance $\sigma_\phi^2(x)$). The reparameterisation $z=\mu_\phi(x)+\sigma_\phi(x)\odot\epsilon$, where $\epsilon\sim\mathcal N(0,I)$, allows gradients to pass through latent draws while the encoder and decoder are trained together. Once trained, unconditional generation draws $z\sim p_Z$ and then $x_{\mathrm{new}}\sim p_\theta(x\mid z)$. 
VAEs thus illustrate how flexible neural latent-variable learning leads to a direct generative procedure through the learned probabilistic decoder.

\subsection{Adversarial learning: generative adversarial networks}
\label{sec:gan}

Adversarial learning takes a different route to fitting a direct generator: instead of evaluating an explicit output likelihood, it improves generated samples through comparison with real observations. In the original generative adversarial network (GAN) \citep{goodfellow2014}, a generator $G_\theta$ maps $z\sim p_Z$ to $x=G_\theta(z)$, while a discriminator $D_\psi(x)\in(0,1)$ learns to distinguish real and generated examples. Although the generator is deterministic for a fixed $z$, random latent inputs induce an output distribution.

The original training criterion is the minimax game, where
$p_{\mathrm{data}}$ denotes the distribution of observed data:
\begin{equation}\label{eq:gan}
\min_\theta\max_\psi\ \Bigl[
\underbrace{\mathbb E_{x\sim p_{\mathrm{data}}}[\log D_\psi(x)]}_{\text{real samples}}
+\underbrace{\mathbb E_{z\sim p_Z}[\log(1-D_\psi(G_\theta(z)))]}_{\text{generated samples}}
\Bigr].
\end{equation}
The discriminator is rewarded for classifying observed samples as real and generated samples as fake. With the discriminator temporarily fixed, its output on $G_\theta(z)$ supplies a differentiable signal for changing the generator's parameters; the two networks are commonly updated alternately. Thus the discriminator is a training device that identifies weaknesses in the current generator, not a step in the final sampling procedure. A frequently used practical alternative for the generator is the non-saturating loss $-\mathbb E_{z\sim p_Z}[\log D_\psi(G_\theta(z))]$, which can provide a stronger learning signal when early samples are easy to distinguish.

This challenge helps motivate the Wasserstein GAN (WGAN) \citep{arjovsky2017}. When real and generated distributions lie on separated supports, divergence-based comparisons associated with the idealised original GAN game can saturate and provide poor gradient information. WGAN instead uses a Lipschitz-constrained, real-valued critic and a learning criterion related to the Wasserstein-1 distance, which can remain informative even for separated distributions. WGAN-GP uses a gradient penalty to encourage the critic's Lipschitz constraint \citep{gulrajani2017}. 

The central generation mechanism is unchanged across these adversarial variants. Training learns $G_\theta$ through a discriminator or critic, but after training a fresh $z\sim p_Z$ is simply passed through $G_\theta$ to obtain a complete observation. The discriminator or critic is not required for generation, and no explicit generated-data likelihood need be evaluated.

\subsection{Invertible transformations: finite-layer normalising flows}
\label{sec:nf}

Normalising flows (NFs) \citep{papamakarios2021} offer a third learning principle: construct a reversible transformation between a simple base distribution and the observation space. In a conventional finite-layer flow, a differentiable bijection $f_\theta$ is built by composing invertible layers. It maps $z$ to $x$ and can also map $x$ back to $z$:
\begin{equation}\label{eq:nf-sample}
 z\sim p_Z(z),\qquad x=f_\theta(z),\qquad z=f_\theta^{-1}(x).
\end{equation}
Unlike a conventional VAE decoder or GAN generator, this construction guarantees an exact inverse under its assumptions.

Reversibility is particularly useful because it connects generation to explicit density evaluation. A change of variables gives
\begin{equation}\label{eq:nf}
 \log p_\theta(x)
 =\underbrace{\log p_Z\!\left(f_\theta^{-1}(x)\right)}_{\text{base-space log-density}}
 +\underbrace{\log\left|\det J_{f_\theta^{-1}}(x)\right|}_{\text{volume correction}}.
\end{equation}
These layers transform the base distribution into a learned observation distribution. The Jacobian determinant accounts for the resulting local volume changes: expanding a region reduces density for the same probability mass, whereas contracting it increases density. This identity makes maximum-likelihood training possible by minimising $\mathcal L_{\mathrm{NF}}(\theta)=-\mathbb E_{x\sim p_{\mathrm{data}}}[\log p_\theta(x)]$. Learning therefore changes the transformation so that the base-space fit and local volume corrections together assign high likelihood to observed examples.

A central challenge is to make the mapping invertible and the Jacobian determinant inexpensive to compute. Planar flows illustrate a low-rank transformation with an efficient determinant, although invertibility conditions and inverse evaluation require care \citep{rezende2015}. An affine coupling layer, as in Real NVP, takes a different approach: it leaves one part $z_a$ unchanged and transforms the other as $y_b=z_b\odot\exp s(z_a)+t(z_a)$, where small networks compute a scale $s$ and shift $t$ \citep{dinh2017}. Given $y_a=z_a$, reversing the scale and shift recovers $z_b$; the forward log-Jacobian determinant is the sum of the scale outputs, while the inverse contributes the negative sum. Stacking layers with different partitions creates a flexible transformation while preserving easy inversion and density correction. Glow also uses invertible $1\times1$ convolutions to mix components between such layers \citep{kingma2018glow}. These designs trade unrestricted architecture for tractable likelihood and sampling.

For a finite-layer NF, generation simply evaluates the learned
composite map forwards: draw $z\sim p_Z$ and compute $x=f_\theta(z)$.
Likelihood evaluation instead uses the inverse map and its Jacobian
correction. Although the forward map passes through multiple layers,
it is evaluated as one specified generator, rather than by a separate
iterative sampling algorithm.

Continuous normalising flows (CNFs) provide a useful boundary
case: they define an invertible transformation through an ordinary
differential equation $dx(t)/dt=v_\theta(x(t),t)$
\citep{chen2018neuralode}. The density along its trajectory obeys
\begin{equation}\label{eq:cnf-density}
 \frac{d}{dt}\log p_t(x(t))
 =-\nabla\!\cdot v_\theta(x(t),t).
\end{equation}
Their numerical integration updates an evolving state, linking CNFs
to the iterative generation route and to flow matching in
Section~\ref{sec:fm}.

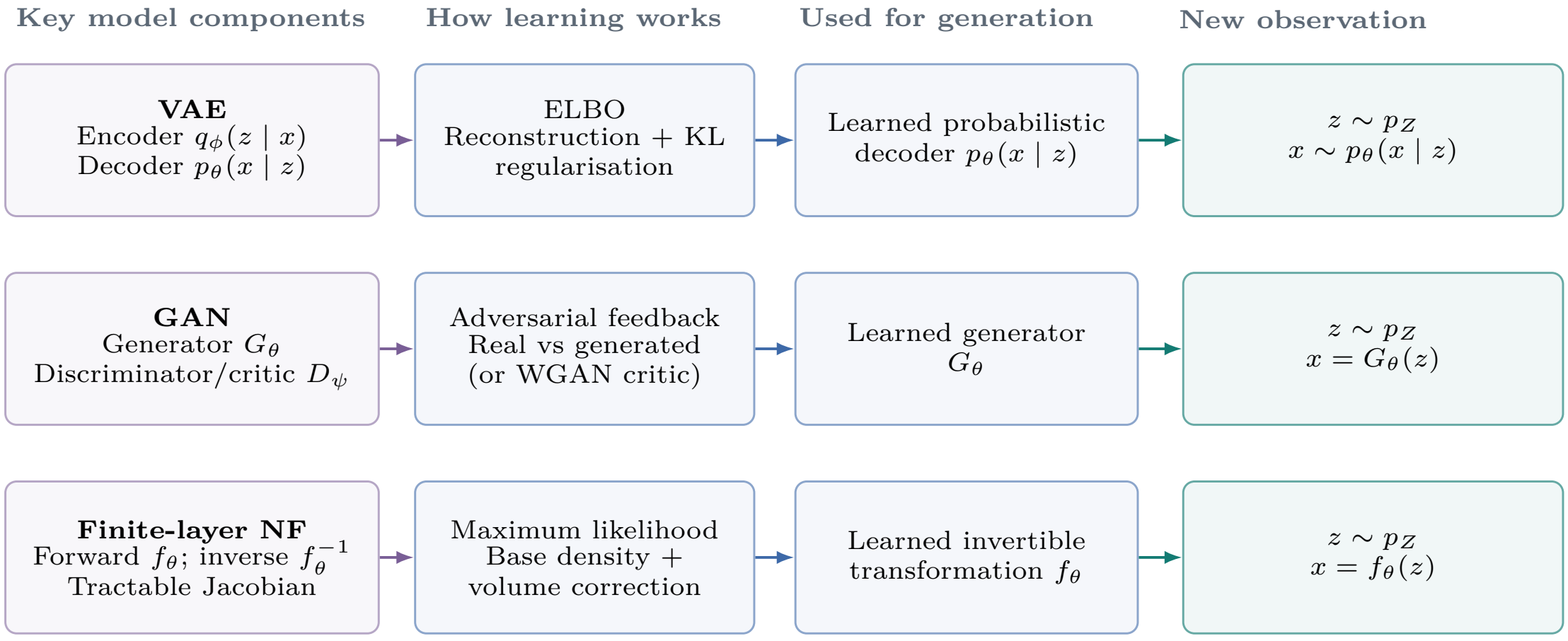
\begin{figure}[H]
 \centering
 % Integrated direct-generation comparison: components, learning and sampling.
% Colour names and arrow styles are defined in main.tex.
\begin{tikzpicture}[font=\scriptsize,>=Latex]
\tikzset{directcell/.style={rounded corners=3pt,draw,minimum height=1.42cm,align=center,inner sep=4pt,line width=.55pt}, directarrow/.style={-{Latex[length=1.65mm]},line width=.72pt}}
\node[font=\bfseries\scriptsize,text=mute,anchor=west] at (0.02,1.13) {Key model components};
\node[font=\bfseries\scriptsize,text=mute,anchor=west] at (3.85,1.13) {How learning works};
\node[font=\bfseries\scriptsize,text=mute,anchor=west] at (7.34,1.13) {Used for generation};
\node[font=\bfseries\scriptsize,text=mute,anchor=west] at (10.89,1.13) {New observation};

\node[directcell,draw=purple!55,fill=purple!5,minimum width=3.49cm] (v1) at (1.79,0) {\textbf{VAE}\\[-1pt] Encoder $q_\phi(z\mid x)$\\ Decoder $p_\theta(x\mid z)$};
\node[directcell,draw=blue!60,fill=blue!5,minimum width=3.17cm] (v2) at (5.46,0) {ELBO\\[-1pt] Reconstruction + KL\\ regularisation};
\node[directcell,draw=blue!60,fill=blue!5,minimum width=3.20cm] (v3) at (9.03,0) {Learned probabilistic\\ decoder $p_\theta(x\mid z)$};
\node[directcell,draw=teal!65,fill=teal!6,minimum width=3.55cm] (v4) at (12.83,0) {$z\sim p_Z$\\ $x\sim p_\theta(x\mid z)$};
\draw[directarrow,purple] (v1.east)--(v2.west);
\draw[directarrow,blue] (v2.east)--(v3.west);
\draw[directarrow,teal] (v3.east)--(v4.west);

\node[directcell,draw=purple!55,fill=purple!5,minimum width=3.49cm] (g1) at (1.79,-1.95) {\textbf{GAN}\\[-1pt] Generator $G_\theta$\\ Discriminator/critic $D_\psi$};
\node[directcell,draw=blue!60,fill=blue!5,minimum width=3.17cm] (g2) at (5.46,-1.95) {Adversarial feedback\\[-1pt] Real vs generated\\ (or WGAN critic)};
\node[directcell,draw=blue!60,fill=blue!5,minimum width=3.20cm] (g3) at (9.03,-1.95) {Learned generator\\ $G_\theta$};
\node[directcell,draw=teal!65,fill=teal!6,minimum width=3.55cm] (g4) at (12.83,-1.95) {$z\sim p_Z$\\ $x=G_\theta(z)$};
\draw[directarrow,purple] (g1.east)--(g2.west);
\draw[directarrow,blue] (g2.east)--(g3.west);
\draw[directarrow,teal] (g3.east)--(g4.west);

\node[directcell,draw=purple!55,fill=purple!5,minimum width=3.49cm] (f1) at (1.79,-3.90) {\textbf{Finite-layer NF}\\[-1pt] Forward $f_\theta$; inverse $f_\theta^{-1}$\\ Tractable Jacobian};
\node[directcell,draw=blue!60,fill=blue!5,minimum width=3.17cm] (f2) at (5.46,-3.90) {Maximum likelihood\\[-1pt] Base density +\\ volume correction};
\node[directcell,draw=blue!60,fill=blue!5,minimum width=3.20cm] (f3) at (9.03,-3.90) {Learned invertible\\ transformation $f_\theta$};
\node[directcell,draw=teal!65,fill=teal!6,minimum width=3.55cm] (f4) at (12.83,-3.90) {$z\sim p_Z$\\ $x=f_\theta(z)$};
\draw[directarrow,purple] (f1.east)--(f2.west);
\draw[directarrow,blue] (f2.east)--(f3.west);
\draw[directarrow,teal] (f3.east)--(f4.west);
\end{tikzpicture}
 \caption{\textbf{Direct generation: latent-variable, adversarial and invertible-map approaches.}}
 \label{fig:direct-comparison}
\end{figure}

Figure~\ref{fig:direct-comparison} makes the key comparison explicit: the three approaches use different components and training signals, yet each can produce a complete observation from a newly sampled input. These are complementary learning perspectives, not mutually exclusive classes: GANs and NFs also use latent variables, and different mechanisms may be combined as discussed in Section~\ref{sec:compare}. This shared generation-time pattern provides the point of comparison with sequential and iterative generation in the following sections.

\section{Sequential generation: constructing observations through conditional draws}
\label{sec:ar}

Sequential generation constructs a complete observation progressively,
rather than producing it directly from a sampled input.
The central idea is to divide an observation into a sequence of
components and generate them one at a time in a chosen order.
At each step, the next component is sampled conditionally on
those already generated, following a procedure such as
$x_i\sim p_\theta(\cdot\mid x_{<i})$,
where $x_{<i}$ denotes the components generated before step $i$.
Each newly generated component becomes part of the context for
subsequent steps, until the complete observation has been formed.

\subsection{Autoregressive factorisation: the common principle}
\label{sec:ar-factor}

The probability chain rule decomposes a joint distribution into
conditional distributions:
\begin{equation}\label{eq:ar}
 p_\theta(x)=\prod_{i=1}^{n}p_\theta(x_i\mid x_{<i}),
 \qquad x_{<i}=(x_1,\ldots,x_{i-1}).
\end{equation}
This factorisation holds for any joint distribution.
An autoregressive model makes it generative by learning
the conditional distributions so that it can subsequently sample
each new component given those already produced.
The first factor is unconditional (or conditional only on
externally supplied context).

Learning commonly minimises the negative log-likelihood of
observed examples:
\begin{equation}\label{eq:ar-loss}
 \mathcal L_{\mathrm{AR}}(\theta)
 =-\mathbb E_{x\sim p_{\mathrm{data}}}\!\left[
 \sum_{i=1}^{n}\log p_\theta(x_i\mid x_{<i})\right].
\end{equation}
This objective trains the model to assign high conditional
probability to each observed component given its preceding
components. For categorical tokens, a neural network outputs
probabilities over possible next tokens, and learning corresponds
to cross-entropy training. Although this resembles a classification
task, the predicted probabilities describe the conditional
distributions needed for generation. Together, these learned
conditionals define a probability distribution over complete
observations.

\begin{figure}[!htbp]
\centering
\begin{tikzpicture}[font=\footnotesize,>=Latex]
\node[font=\bfseries,anchor=west,text=purple] at (0,1.63) {(a) Learning: observed prefixes and observed targets};
\node[nodebox,cond,minimum width=2.5cm] (ta) at (1.73,.81) {Observed prefix $x_{<i}$};
\node[nodebox,learn,minimum width=3.18cm] (tb) at (6.00,.81) {Conditional $p_\theta(x_i\mid x_{<i})$};
\node[nodebox,learn,minimum width=2.68cm] (tc) at (10.69,.81) {Loss against\\observed $x_i$};
\draw[arrow,purple] (ta)--(tb);\draw[arrow,purple] (tb)--(tc);
\node[font=\scriptsize,text=mute,anchor=west] at (.48,.17) {Likelihood training rewards probability assigned to the actual next component.};
\draw[black!20] (0,-.16)--(13.38,-.16);
\node[font=\bfseries,anchor=west,text=blue] at (0,-.48) {(b) Generation: sample conditionally and extend the prefix};
\node[nodebox,cond,minimum width=2.35cm] (ga) at (1.66,-1.32) {Current prefix};
\node[nodebox,cond,minimum width=3.19cm] (gb) at (5.75,-1.32) {Learned conditional\\$p_\theta(x_i\mid x_{<i})$};
\node[nodebox,sample,minimum width=2.16cm] (gc) at (9.88,-1.32) {Draw $x_i$};
\draw[arrow,blue] (ga)--(gb);\draw[arrow,blue] (gb)--(gc);
\draw[dottedarrow,blue] (gc.south) -- ++(0,-.40) -| node[pos=.29,below,font=\scriptsize] {append to prefix; repeat} (ga.south);
\end{tikzpicture}
\caption{\textbf{Autoregressive learning and sequential generation.}}
\label{fig:ar}
\end{figure}
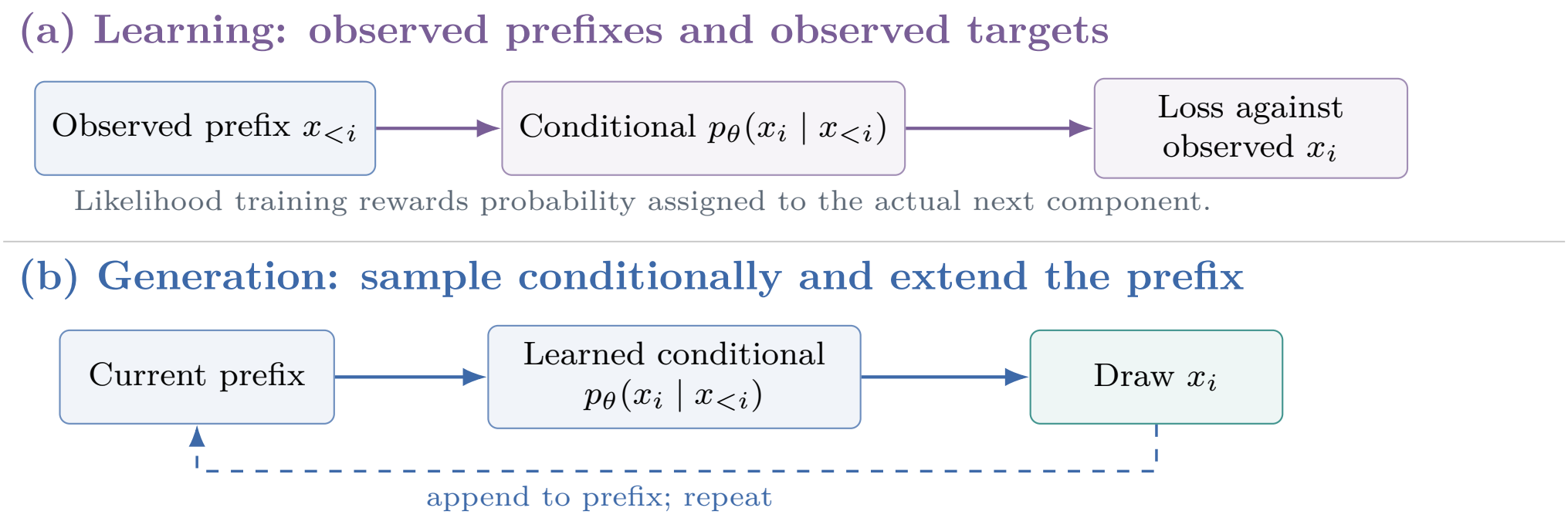

During training, the model uses observed prefixes to predict
their next components, a practice often called teacher forcing \citep{williams1989}.
During generation, the next conditional is instead evaluated using
previously generated components: draw $x_1\sim p_\theta(x_1)$,
then $x_2\sim p_\theta(\cdot\mid x_1)$, and continue until the
observation is complete. Thus, the conditional distributions
learned from observed data become the sampling rules for
generating new observations. Figure~\ref{fig:ar} illustrates
this distinction between learning from observed prefixes
and sampling from generated ones.

\subsection{Different architectures, one conditional generation mechanism}
\label{sec:ar-architectures}

Autoregression specifies a probabilistic modelling principle,
rather than a particular neural architecture. Given the conditional
factorisation and learning objective described above, a central
implementation challenge is how to represent the preceding
components and predict the next-component distribution without
accessing future information. Different neural architectures
address this challenge in different ways, while following the
same underlying generation principle.

Recurrent neural networks (RNNs) and long short-term memory
networks (LSTMs) provide early examples. Although commonly used
for classification and regression, these architectures can also
support generative modelling \citep{mikolov2010}. In an autoregressive RNN, a hidden
state carries information from previously processed components,
and an output layer uses this state to predict the next-component
distribution. LSTMs introduce gates to better retain relevant
information across longer sequences \citep{hochreiter1997}.
During generation, each newly sampled component is fed back
into the network to predict the next one.

Other architectures were developed more explicitly for
autoregressive generation. PixelCNN, for example, models
images through conditional distributions over successive pixels.
It uses masked convolutions to prevent each prediction from
accessing pixels that come later in the chosen ordering, while
stacking layers allows information from a wider preceding region
to be used \citep{oord2016pixel}. A related approach,
WaveNet, applies causal convolutions to raw audio generation.
Its dilated convolutions allow each prediction to incorporate
information from many preceding audio samples, while generation
proceeds by sampling successive waveform values
\citep{oord2016wavenet}.

A causal Transformer instead uses masked self-attention
to represent relationships among preceding tokens. Unlike an
RNN's running hidden state, attention allows each position
to combine information from multiple earlier positions directly.
Causal masking prevents access to future tokens, and the resulting
representation is used to predict a distribution over the next
token \citep{vaswani2017,radford2019}.

These architectures differ in how they represent preceding
information and compute conditional distributions, but they
support the same autoregressive generation principle. During
training, masked convolutional and Transformer architectures
can compute predictions at many positions in parallel using
observed prefixes. During generation, however, each new
component depends on previously sampled components, so ordinary
autoregressive sampling proceeds sequentially.

\section{Iterative generation: updating a candidate using learned guidance}
\label{sec:iter}

Whereas sequential generation constructs an observation component
by component, iterative generation starts with an initial state
and repeatedly updates it to produce a sample reflecting the
statistical structure learned from observations. An update can
modify part or all of the existing state. Schematically,
$x^{(k+1)}=F_{\theta,k}(x^{(k)})$, where $F_{\theta,k}$
represents a sampling-time update. Three influential approaches
differ primarily in the guidance they learn: denoising or score
information, velocity fields and energy functions.

\subsection{Diffusion models: denoising and score-based perspectives}
\label{sec:diff}

Diffusion models have been central to recent advances in image
generation, showing how complex observations can emerge from a
sequence of learned updates. Two closely related perspectives help
explain the mechanism: probabilistic denoising, as in denoising
diffusion probabilistic models (DDPMs) \citep{sohldickstein2015,ho2020}, and score-based generation \citep{song2019score}.
Both start from noise and progressively transform an existing
state towards the distribution of observations.

A DDPM describes two processes. Its prescribed forward process
gradually corrupts observed data with Gaussian noise until the
resulting distribution is approximately a simple Gaussian. The
learned reverse process starts from that noise distribution
and samples a sequence of less noisy states to produce a new
observation. In the standard construction, a noisy
state at level $t$ can be obtained directly from $x_0$:
\begin{equation}\label{eq:noising}
 x_t=\sqrt{\bar\alpha_t}\,x_0+
 \sqrt{1-\bar\alpha_t}\,\epsilon,
 \qquad \epsilon\sim\mathcal N(0,I),
\end{equation}
where $\bar\alpha_t$ is determined by the noise schedule. As the
noise level increases, the observation's structure becomes harder
to discern. Since the injected noise is known during training,
the model can learn to predict it from $x_t$ and $t$.

DDPMs also admit a probabilistic latent-variable interpretation:
the intermediate noisy states serve as latent variables, and the
reverse model can be learned through a variational objective,
connecting to Section~\ref{sec:latent-direct}. A widely used
simplified training objective is noise-prediction error:
\begin{equation}\label{eq:ddpm-loss}
 \mathcal L_{\mathrm{noise}}(\theta)=
 \E_{x_0,t,\epsilon}
 \bigl[\|\epsilon-\epsilon_\theta(x_t,t)\|_2^2\bigr].
\end{equation}
Predicting noise at different levels teaches the network how
corrupted observations relate to the data. In particular,
rearranging the noising equation gives an estimate of the clean
observation:
\begin{equation}\label{eq:ddpm-x0}
 \widehat x_0(x_t,t)=
 \frac{x_t-\sqrt{1-\bar\alpha_t}\,
 \epsilon_\theta(x_t,t)}{\sqrt{\bar\alpha_t}}.
\end{equation}
This estimate helps determine the next reverse transition; it is
not itself the entire sampling step. During generation, the
predicted noise, known schedule and reverse-transition variance
are used to sample $x_{t-1}$ from $x_t$, repeatedly updating the
current state until a new observation is obtained.

The score-based perspective makes the guidance behind such
updates more explicit. For a density $p(x)$, its score
$s(x)=\nabla_x\log p(x)$ points locally towards increasing density.
A learned score can therefore guide a sample's movement. For
example, Langevin sampling combines this direction with random
perturbations:
\begin{equation}\label{eq:score-langevin}
 x^{(k+1)}=x^{(k)}
 +\frac{\eta}{2}s_\theta(x^{(k)},t)
 +\sqrt{\eta}\,\xi_k,
 \qquad \xi_k\sim\mathcal N(0,I).
\end{equation}
Here $s_\theta(x,t)$ estimates the score at a fixed noise level
$t$. The random term allows exploration rather than simply
moving towards a density maximum.

Direct score estimation can be difficult for complex data.
Adding Gaussian noise creates smoother distributions $p_t(x)$
whose scores provide broader guidance at high noise levels
and finer detail at low levels. Although these marginal scores
are unknown, the conditional Gaussian corruption is known.
This makes it possible to learn a noise-conditioned score using
denoising score matching \citep{vincent2011,song2019score}:
\begin{equation}\label{eq:score-loss}
 \mathcal L_{\mathrm{score}}(\theta)
 =\E_{x_0,t,\epsilon}\!\left[
 \lambda(t)\left\|
 s_\theta(x_t,t)+
 \frac{\epsilon}{\sqrt{1-\bar\alpha_t}}
 \right\|_2^2\right].
\end{equation}
The training target is the known conditional score; under
suitable conditions, its squared-error optimum is the score
of the noisy marginal $p_t(x)$.

During generation, annealed Langevin sampling uses scores
at successively lower noise levels \citep{song2019score}. Related approaches employ
reverse-time stochastic dynamics or a probability-flow
ordinary differential equation (ODE), solved through numerical
time steps \citep{song2021}. These procedures differ, but share
the use of learned guidance to evolve a sample.

Indeed, under the same Gaussian corruption, DDPM noise prediction
and score estimation are connected by
$s_\theta(x_t,t)\approx
-\epsilon_\theta(x_t,t)/\sqrt{1-\bar\alpha_t}$.
Thus noise prediction also supplies information about the score
of the noisy data distribution. The two perspectives explain
related ways to use learned statistical structure for generation:
in each case, learning guides repeated updates of a
current state, rather than extending an observation component
by component.

\subsection{Velocity-based transport: flow matching}
\label{sec:fm}

Rather than learning scores to guide sampling, flow matching
asks a different question: how should samples move so that a simple
source distribution gradually transforms into the data distribution?
The central idea is to learn a time-dependent velocity field
$v_\theta(x,t)$ that describes how a sample should move at each
position and time, without explicitly learning a normalised density
\citep{lipman2023}.

To construct a learning objective, consider a simple example.
Draw $z\sim p_Z$ and $y\sim p_{\mathrm{data}}$ independently,
and connect them through a straight path
$x_t=(1-t)z+ty$, where $t\in[0,1]$.
The velocity along this path, obtained by differentiating
$x_t$ with respect to time, is simply $y-z$.
This gives the regression objective
\begin{equation}\label{eq:fm-loss}
 \mathcal L_{\mathrm{FM}}(\theta)=\E_{t,z,y}
 \bigl[\|v_\theta(x_t,t)-(y-z)\|_2^2\bigr].
\end{equation}
Here, each sampled pair provides an intermediate position
$x_t$ and its associated velocity $y-z$.
However, different endpoint pairs may pass through the same
position at the same time, with different velocities.
Squared-error regression therefore learns their conditional
mean velocity, $\E[y-z\mid x_t=x]$, which, under suitable
conditions, defines a velocity field consistent with the
evolving distribution of the interpolated points.
The network thus learns which velocity arrow to assign
at a given position and time, rather than memorising
individual paths or predicting a particular endpoint.
This explains how independently paired source and data samples
can be used to learn a shared transport field.
Other flow-matching formulations may use different
couplings or paths  \citep{lipman2023}.

Once this velocity field has been learned, generation begins
with $x(0)\sim p_Z$ and evolves the sample according to
the ordinary differential equation
$dx(t)/dt=v_\theta(x(t),t)$, for $t\in[0,1]$.
In practice, a numerical solver repeatedly evaluates the
learned velocity at the current position and time to
determine the next state. For example, an Euler update
takes the form
\begin{equation}\label{eq:fm-euler}
 x^{(k+1)}=x^{(k)}+\Delta t_k\,v_\theta(x^{(k)},t_k).
\end{equation}
Following these updates transports the initial sample
along a learned trajectory towards the data distribution.
More accurate numerical solvers can use the same velocity field.

Under suitable regularity conditions, the ODE defines an
invertible transformation from the initial state $x(0)$
to the final state $x(1)$. This is precisely the mechanism
of a continuous normalising flow \citep{chen2018neuralode}: unlike finite-layer
normalising flows, whose transformations are evaluated through
a fixed composition of invertible layers, the transformation
is obtained by integrating a velocity field over time.
Flow matching provides a way to learn this field,
rather than requiring likelihood-based training of the flow.
Score-based probability-flow ODEs also describe continuous
sample transport, but obtain their dynamics through a
different learning principle.

\subsection{Energy-guided sampling: energy-based models}
\label{sec:ebm}

Energy-based models (EBMs) take a different perspective,
drawing on the concept of energy from statistical physics \citep{lecun2007}.
The central idea is that configurations with lower energy are
more favoured than those with higher energy. Through a
Boltzmann-type distribution, a learned scalar energy
$E_\theta(x)$ defines a probability density:
\begin{equation}\label{eq:energy}
 p_\theta(x)=\frac{e^{-E_\theta(x)}}{Z_\theta},\qquad
 Z_\theta=\int e^{-E_\theta(u)}\,du.
\end{equation}
Here, $Z_\theta$, known as the partition function,
normalises the distribution. Rather than learning a direct
mapping from noise to observations, the model learns an energy
landscape describing which configurations are relatively
favoured. However, this landscape does not by itself provide
a direct procedure for drawing new observations.

A representative learning approach is maximum likelihood,
with the objective
\begin{equation}\label{eq:ebm-loss}
 \mathcal L_{\mathrm{EBM}}(\theta)
 =\E_{p_{\mathrm{data}}}[E_\theta(x)]+\log Z_\theta.
\end{equation}
Under standard regularity assumptions, its gradient is
\begin{equation}\label{eq:ebm-gradient}
 \nabla_\theta\mathcal L_{\mathrm{EBM}}
 =\E_{p_{\mathrm{data}}}[\nabla_\theta E_\theta(x)]
 -\E_{p_\theta}[\nabla_\theta E_\theta(x)].
\end{equation}
The two expectations have complementary roles: learning
encourages lower energy for observed configurations while
discouraging the model from assigning low energy indiscriminately
elsewhere. Estimating the second expectation, often called the negative phase, may require samples from the current
model distribution. These can be obtained through Markov-chain
methods such as the Langevin update in
Eq.~\eqref{eq:score-langevin} \citep{grathwohl2020}.
Here, sampling is part of training: the generated
configurations help estimate the gradient used to update
the energy model.

After training, sampling serves a different purpose: producing
new observations from the fitted model while keeping its
parameters fixed. For an EBM, the score of its density is
$\nabla_x\log p_\theta(x)=-\nabla_x E_\theta(x)$,
since the partition function does not depend on $x$.
The same Langevin update can therefore be used with the
negative energy gradient guiding the movement of a current
sample. Repeated updates can approximate samples from
$p_\theta$, although discretisation and inadequate mixing
may affect their quality.

This reveals a connection with score-based generation through
a shared sampling mechanism, despite their different learning
principles. A conventional EBM learns an energy function
defining a distribution, whereas diffusion and score-based
models learn noise-dependent denoising or score information.
In either case, the learned information can guide
repeated updates of an existing state, progressively
producing samples according to the modelled statistical
structure rather than constructing an observation
component by component.

\begin{figure}[H]
\centering
% Learning-to-sampling connection for three iterative families.
\begin{tikzpicture}[x=1cm,y=1cm,>=Latex,font=\sffamily]
\node[anchor=west,font=\bfseries\scriptsize,text=mute] at (.06,5.20) {WHAT TRAINING PROVIDES};
\node[anchor=west,font=\bfseries\scriptsize,text=mute] at (5.05,5.20) {HOW THE SAMPLER USES IT};
\node[anchor=west,font=\bfseries\scriptsize,text=mute] at (10.53,5.20) {REPEATED EVOLUTION OF A STATE};
\foreach \y/\col in {3.52/ochre,1.89/blue,.26/purple}{
 \path[fill=\col!5,draw=\col!37,rounded corners=4pt] (.02,\y) rectangle ++(15.45,1.49);
}
% first row
\node[anchor=west,font=\bfseries\footnotesize,text=ochre] at (.19,4.65) {Diffusion / score models};
\node[anchor=west,font=\footnotesize,text width=4.15cm,align=left] at (.19,4.20) {Denoising / score losses\\$\Rightarrow$ noise or score estimate};
\node[anchor=west,font=\footnotesize,align=left,text width=4.8cm] at (5.05,4.34) {Noise prediction guides denoising;\\scores guide Langevin or ODE steps};
\node[anchor=west,font=\footnotesize,text=ochre] at (10.65,4.10) {$x_T\;\longrightarrow\;x_{t}\;\longrightarrow\;x_0$};
\node[anchor=west,font=\footnotesize,text=mute] at (10.65,3.78) {noisy $\rightarrow$ structured};
% second row
\node[anchor=west,font=\bfseries\footnotesize,text=blue] at (.19,3.02) {Flow matching};
\node[anchor=west,font=\footnotesize,text width=4.15cm,align=left] at (.19,2.57) {Velocity-regression objective\\$\Rightarrow$ time-dependent field $v_\theta(x,t)$};
\node[anchor=west,font=\footnotesize,align=left,text width=4.8cm] at (5.05,2.71) {ODE solver queries the field\\and advances the sample};
\node[anchor=west,font=\footnotesize,text=blue] at (10.65,2.47) {$x(0)\;\longrightarrow\;x(t)\;\longrightarrow\;x(1)$};
\node[anchor=west,font=\footnotesize,text=mute] at (10.65,2.15) {source $\rightarrow$ transported sample};
% third row
\node[anchor=west,font=\bfseries\footnotesize,text=purple] at (.19,1.39) {Energy-based sampling};
\node[anchor=west,font=\footnotesize,text width=4.15cm,align=left] at (.19,.94) {Likelihood / contrastive loss\\$\Rightarrow$ energy $E_\theta(x)$};
\node[anchor=west,font=\footnotesize,align=left,text width=4.8cm] at (5.05,1.08) {Langevin MCMC combines $-\nabla E_\theta$\\with random perturbations};
\node[anchor=west,font=\footnotesize,text=purple] at (10.65,.84) {$x^{(0)}\;\longrightarrow\;x^{(1)}\;\longrightarrow\;x^{(K)}$};
\node[anchor=west,font=\footnotesize,text=mute] at (10.65,.52) {after chain exploration};
\end{tikzpicture}
\caption{\textbf{Iterative generation: learned guidance and sampling updates across three approaches.}}\label{fig:iterative}
\end{figure}
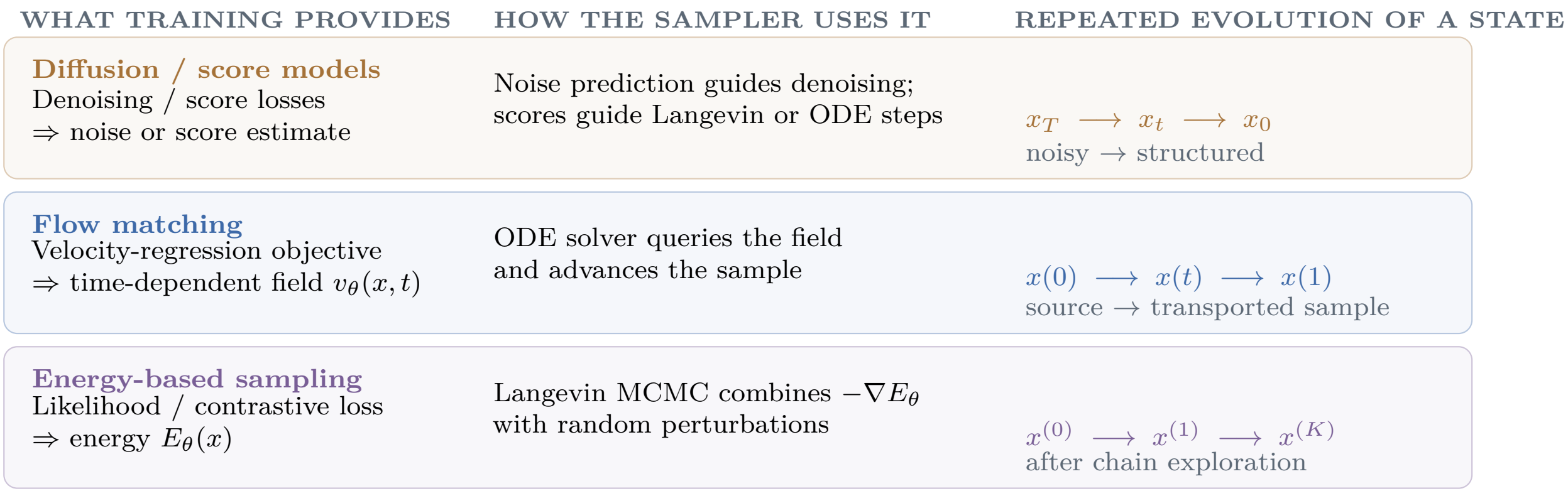

Figure~\ref{fig:iterative} contrasts the learned guidance and repeated sampling operations used by these three iterative approaches.

\section{Connecting and combining generation mechanisms}
\label{sec:compare}

The three generation routes provide a practical starting point for
understanding how new observations are produced. A second view
asks what information each model learns and how that information
is used at sampling time. Bringing these views together also helps
explain systems that cross or combine the routes.

\subsection{What learning supplies and generation requires}

Table~\ref{tab:big} compares representative learning principles,
learned quantities and generation procedures. It is not a ranking
of methods, nor does it assume that every method learns an
explicitly evaluable probability density. Several connections emerge from this comparison. 

\begin{table}[!htbp]
\centering\footnotesize\renewcommand{\arraystretch}{1.19}
\caption{\textbf{Representative modelling principles, learned quantities and generation procedures.}}\label{tab:big}
\begin{tabularx}{\linewidth}{@{}>{\raggedright\arraybackslash}p{1.93cm}>{\raggedright\arraybackslash}p{3.0cm}>{\raggedright\arraybackslash}p{3.42cm}X@{}}
\toprule
\textbf{Family (route)} & \textbf{What is learned?} & \textbf{What does training encourage?} & \textbf{How does it generate?}\\
\midrule
VAE (direct) & Decoder law $p_\theta(x\mid z)$; approximate encoder & ELBO: fit observations under inferred latents while controlling departure from the prior & Draw $z$ from the prior, then sample from the decoder law\\
GAN (direct) & Generator $G_\theta$; discriminator used in training & Adversarial feedback: make generated outputs harder to distinguish from real observations & Draw $z\sim p_Z$, then evaluate $G_\theta(z)$\\
Finite-layer normalising flow (direct) &
Invertible mapping $f_\theta$ and its density transformation &
Likelihood: fit observations through the change-of-variables formula &
Draw $z\sim p_Z$ and evaluate $f_\theta(z)$\\
Autoregressive (sequential) & Conditional laws $p_\theta(x_i\mid x_{<i})$ & Conditional likelihood: predict each observed component from the preceding ones & Sample and append each component given the generated prefix\\
Diffusion / score (iterative) & Denoiser, noise predictor or score-related field & Denoising or score-matching loss: recover appropriate information at noisy states & Repeatedly apply learned reverse-time or refinement updates\\
Flow matching (iterative transport) & Time-dependent velocity $v_\theta(x,t)$ & Velocity regression: approximate motion along a chosen probability path & Sample a base point and integrate the fitted ODE\\
Energy-based (iterative sampling) & Energy $E_\theta(x)$ specifying relative preference & Likelihood or contrastive criteria: favour data relative to model configurations & Run an appropriate sampler, often MCMC, using the energy or its gradient\\
\bottomrule
\end{tabularx}
\end{table}

First,
components used during learning need not participate in
generation. A VAE encoder infers latent explanations for
observed data during training, but new samples use its prior
and decoder. Similarly, a GAN discriminator or critic guides
training but is discarded at generation time. In contrast,
a finite-layer normalising flow uses its reversible map for
both density evaluation and sampling. A shared generation
route can therefore conceal different learning principles.

Second, the learned quantity does not always specify
a complete sampling algorithm. Autoregressive conditionals
must be evaluated as the generated prefix grows; diffusion
noise predictions and scores guide reverse-time updates;
flow-matching velocities require numerical integration; and
energy functions generally require an additional sampler.
Consequently, the implemented output distribution can also
depend on numerical discretisation, stochastic transitions
or the mixing of a Markov chain \citep{song2021,grathwohl2020}.

Another useful distinction is where randomness enters
generation. A GAN or finite-layer flow typically draws a random
input before applying its learned mapping, whereas autoregressive
models draw successive components from conditional laws.
Stochastic diffusion samplers and Langevin chains also inject
randomness during their updates. By contrast, an ODE-based
transport sampler follows a deterministic trajectory once its
random initial state is fixed. The generation route and the
placement of randomness are therefore related but distinct
characteristics. Conditioning on labels, prompts or other context
can likewise be incorporated across different routes \citep{mirza2014conditional,radford2019,ho2022classifierfree}.

These differences have practical consequences. Evaluating a
direct generator can yield a complete observation without a
separate sampling trajectory, whereas sequential methods
make successive conditional draws and iterative methods
typically apply repeated updates. These are tendencies rather
than fixed efficiency rankings: model designs and sampling
procedures can change the computational trade-offs.

\subsection{Routes in combination: composition, nesting and overlap}

\begin{figure}[!ht]
\centering
\begin{tikzpicture}[x=1cm,y=1cm,font=\footnotesize]
\node[font=\bfseries,text=ochre,anchor=west] at (0,2.1) {Latent diffusion: routes composed in sequence};
\node[nodebox,evolve,minimum width=1.9cm] (la) at (1.0,1.55) {latent noise};
\node[nodebox,evolve,minimum width=2.1cm] (lb) at (4.0,1.55) {iterative latent\\denoising};
\node[nodebox,sample,minimum width=1.8cm] (lc) at (7.15,1.55) {decoder};
\node[nodebox,sample,minimum width=1.6cm] (ld) at (10.05,1.55) {image};
\draw[arrow,ochre] (la)--(lb);\draw[arrow,teal] (lb)--(lc);\draw[arrow,teal] (lc)--(ld);
\draw[thin,black!15] (0,.85)--(11.75,.85);
\node[font=\bfseries,text=blue,anchor=west] at (0,.49) {Conditional diffusion: routes nested at two levels};
\node[nodebox,cond,minimum width=2.1cm] (ma) at (1.10,-.44) {generated prefix};
\node[nodebox,cond,minimum width=2.75cm] (mb) at (4.5,-.44) {autoregressive\\conditioning\\representation};
\node[nodebox,evolve,minimum width=2.5cm] (mc) at (8.04,-.44) {iterative conditional\\sampler};
\node[nodebox,cond,minimum width=1.45cm] (md) at (10.95,-.44) {next token};
\draw[arrow,blue] (ma)--(mb);\draw[arrow,ochre] (mb)--(mc);\draw[arrow,blue] (mc)--(md);
\draw[arrow,blue,dashed] (md.south) -- ++(0,-.30) -| node[pos=.25,below,font=\scriptsize] {extend prefix; repeat} (ma.south);
\end{tikzpicture}
\caption{\textbf{Combining routes: latent diffusion and autoregressive conditional diffusion.}}\label{fig:hybrid}
\end{figure}
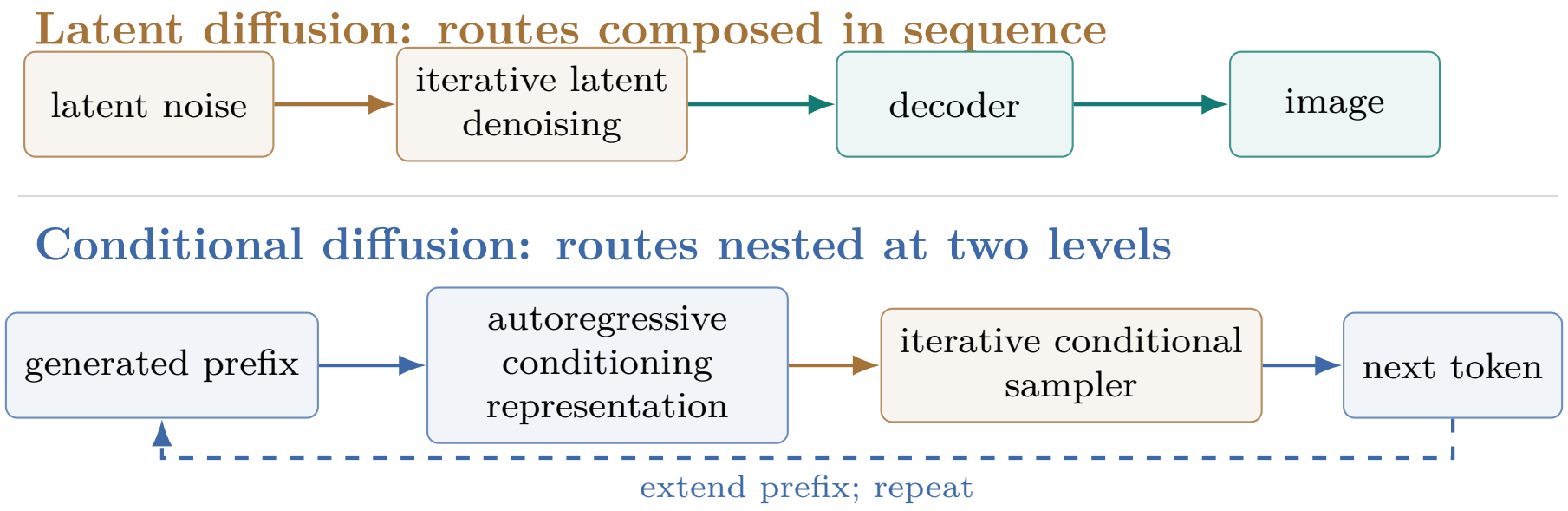

The routes describe generation procedures, not mutually exclusive
model families. In latent diffusion, a diffusion model
iteratively generates a latent representation that is then
decoded into an image \citep{rombach2022}. This combines
iterative latent generation with direct decoding. Alternatively,
a model can construct a sequence autoregressively while using
diffusion to sample each continuous-valued token conditioned on
the generated prefix, as in MAR \citep{li2024mar}. Here a
sequential procedure contains an iterative conditional sampler.
Figure~\ref{fig:hybrid} illustrates these two arrangements.

An architecture likewise does not prescribe a unique route.
For example, a Transformer may support autoregressive token
prediction \citep{radford2019}, iterative masked-token filling
\citep{chang2022maskgit}, or diffusion-based denoising
\citep{peebles2023dit}. The architecture describes how
information is processed; the generation route describes
how the sample is produced. This distinction also applies to
continuous normalising flows, whose learned invertible
transformation is realised through numerical integration,
connecting direct mappings with iterative transport.

Finally, the number of sampling steps is not dictated
solely by a method's training lineage. Consistency models
support one-step or few-step generation \citep{song2023consistency},
while MeanFlow learns an average-velocity-related quantity
for one-step generation \citep{geng2025meanflow}, rather than
merely using a coarse conventional flow-matching solver.
These examples reinforce the usefulness of asking two
questions of any unfamiliar system: what has been learned,
and what operations actually turn it into a sample?

\section{Conclusion}\label{sec:conclusion}
The central question of this paper is operational: how does learning make it possible to produce new observations? We have used direct, sequential and iterative generation to organise the answer, then connected each route to learning objectives, learned quantities and the operations that produce samples. The selected examples are intended to convey enough of each
method's workings to provide a foundation for engaging with more specialised literature, while making connections
across model families apparent.

The resulting picture is deliberately layered rather than exclusive. Similar sampling operations can be supported by different learning principles; a network architecture need not imply a particular generation route; and complete systems can combine routes through composition or nesting. In this sense the framework complements broad surveys and specialist tutorials: it offers a concise technical bridge between understanding individual generative mechanisms and navigating their growing wider landscape.

\vspace{3pt}
\begingroup
\fontsize{7.7pt}{8.6pt}\selectfont
\setlength{\bibsep}{0pt}
\bibliographystyle{concisenat}
\bibliography{references}
\endgroup
\end{document}